\documentclass{article}
\usepackage{ijcai20}

\usepackage{times}
\usepackage{soul}
\usepackage{url}
\usepackage{hyperref}
\usepackage[utf8]{inputenc}
\usepackage[small]{caption}
\usepackage{graphicx}
\usepackage{amsmath}
\usepackage{amsthm}
\usepackage{booktabs}
\usepackage{algorithm}
\usepackage{algorithmic}
\usepackage{subcaption}
\usepackage{amsfonts}
\title{Predicting Landslides Using Locally Aligned Convolutional Neural Networks}

\author{
Ainaz Hajimoradlou$^1$\and
Gioachino Roberti$^2$\and
David Poole$^1$
\affiliations
$^1$University of British Columbia\\
$^2$Minerva Intelligence\\
\emails
\{ainaz, poole\}@cs.ubc.ca,
groberti@minervaintelligence.com
}

\begin{document}

\maketitle

\begin{abstract}
Landslides, movement of soil and rock under the influence of gravity, are common phenomena that cause significant human and economic losses every year. Experts use heterogeneous features such as slope, elevation, land cover, lithology, rock age, and rock family to predict landslides.
To work with such features, we adapted convolutional neural networks to consider relative spatial information for the prediction task. Traditional filters in these networks either have a fixed orientation or are rotationally invariant. Intuitively, the filters should orient uphill, but there is not enough data to learn the concept of uphill; instead, it can be provided as prior knowledge. We propose a model called Locally Aligned Convolutional Neural Network, LACNN, that follows the ground surface at multiple scales to predict possible landslide occurrence for a single point. To validate our method, we created a standardized dataset of georeferenced images consisting of the heterogeneous features as inputs, and compared our method to several baselines, including linear regression, a neural network, and a convolutional network, using log-likelihood error and Receiver Operating Characteristic curves on the test set. Our model achieves 2-7\% improvement in terms of accuracy and 2-15\% boost in terms of log likelihood compared to the other proposed baselines.
\end{abstract}

\section{Introduction}

Landslides, the downslope movement of Earth materials under the influence of gravity, are common and destructive phenomena. Despite the number of studies focusing on landslide mapping~\cite{guzzetti} and landslide spatial and temporal probability prediction~\cite{Reichenbach2018,safeland}, effective real-world predictive models are scarce and landslides cause significant life and economic losses every year~\cite{Petely}. There are three different approaches to landslide susceptibility mapping: expert-based, physical-based, and statistical approaches. Expert-based methods rely on the qualitative judgment of a domain expert, while physical-based approaches model the stability of a slope given physical parameters such as geotechnical rock and soil properties, and calculate the equilibrium between destabilizing factors and slope strength, but often require more information than is available at scale. Statistical models rely on the statistical analysis of large landslide databases and their relation with landscape attributes. Landscape attributes typically include internal (e.g. slope angle, rock type, etc.) and external (e.g. rainfall) properties of the slope. These data are then used to map the spatial and/or temporal probability of slope failure~\cite{safeland}. The spatial probability of landslide occurrence is usually referred to as a susceptibility map. When the magnitude and the temporal component (e.g. frequency and triggers) are also considered, it is referred to as a hazard map~\cite{safeland}.

Statistical approaches for predicting landslides have significantly increased in recent years. However, they mostly apply models such as linear logistic regression, Support Vector Machines (SVM), or neural networks~\cite{Reichenbach2018}. In this study, we propose a novel convolutional model which we call a Locally Aligned Convolutional Neural Network, LACNN, for producing susceptibility maps. Convolutional Neural Networks, CNNs, form a category of neural network models with tied parameters~\cite{CNN}. CNNs with pooling layers can capture both local and global features of an image, which has been proven extremely useful in many vision tasks such as object recognition, image classification, and object detection.

We are interested in predicting the landslide probability for each point on the ground. The output of our model is a probability map with the same resolution as the input features. We use a fully convolutional model for this purpose. These models have been widely used for image segmentation~\cite{FCN,UNet,FCN_1} and usually consist of down-sampling and up-sampling stages. One of the popular models in this category is UNet~\cite{UNet}, which our architecture is based on. The down-sampling stage consists of convolutions with pooling layers and tries to create a set of compact features capturing both local and global properties of the input features. The up-sampling stage typically consists of convolution transpose layers which are mainly doing the inverse of pooling but with learned parameters. We do not use convolution transpose layers in our model as they tended to produce checkboard artifacts in our experiments, which is a common problem in the literature as well~\cite{checkboard}. Instead, we use interpolation for up-sampling. It has been shown that adding skip connections to a fully convolutional model improves its performance~\cite{skip1,skip2}. As short skip connections have been shown to work only in very deep networks, we only apply long connections to our model.

To produce good susceptibility maps for landslides, we proposed learning filters that can follow the ground surface and extract features towards the uphill direction. For this to work, we need the CNN model to preserve orientational information of landslides to each other but this is not possible using traditional techniques, when the filters are either rotationally invariant or align themselves up, down, left, and right, which corresponds to north, south, east, and west.
We add a pre-processing stage to our CNN model to find the best directions for each pixel at multiple scales and then learn hidden features according to those directions. We call this model a Locally Aligned CNN as the model first aligns itself to a specific set of orientations and then learns a classifier.

The contributions of our paper are:
\begin{itemize}
    \item We provide a standardized dataset so that others can compare their results to ours. This dataset is compiled from public domain data from various sources, including the CORINE land cover inventory\footnote{\url{https://land.copernicus.eu/pan-european/corine-land-cover}}, Italian National Geoportal website\footnote{\url{http://www.pcn.minambiente.it/mattm/en/wfs-service/}}, and the National Institute of Geophysics and Volcanology\footnote{\url{http://tinitaly.pi.ingv.it/}}.
    The dataset consists of several input features such as the slope, elevation, rock types with age and family, and land cover, along with the ground truth in the shape of landslide polygons, which can be used in both a supervised and unsupervised learning framework.
    \item We propose a novel statistical approach for predicting landslides using deep convolutional networks. We develop a model that can capture each pixel's orientation at multiple different ranges to classify a landslide. We use ranges of 30, 100, and 300 meters in our model. These scales can be optimized using cross-validation.
    \item We define several baseline models that are representative of the current state of the art for comparison. We provide five different baselines including a Naive model, a linear logistic regression (LLR), a neural network (NN), and a locally aligned neural network (LANN) model without any convolutions to compare our model's performance against them. These baselines can also be seen as ablation studies for our model.
    \item We provide a way to use CNN models with heterogeneous datasets for predicting landslides rather than only using images in our models.
\end{itemize}

\section{Related Work}

Producing susceptibility maps by statistical approaches is not new in the landslide community. Many people have been using models such as logistic regression, SVM, and random forests. Catani et al.~\shortcite{catani2013} used random forests to generate susceptibility maps emphasizing on sensitivity and scaling issues. Micheletti et al.~\shortcite{Micheletti} and Youssef et al.~\shortcite{Youssef} also used random forest models in predicting landslides for Switzerland and Wadi Tayyah Basin in Saudi Arabia. Some have developed software packages using random forests for susceptibility mapping~\cite{Behnia}. Micheletti et al.~\shortcite{Micheletti} generate several susceptibility mappings using SVMs, random forests, and Adaboost. Atkinson and Massari~\shortcite{Atkinson}, Ayalew and Yamagishi~\shortcite{Ayalew}, and Davis et al.~\shortcite{Davis} focus on linear regression for predicting landslides due to its simplicity and easy training procedure. There is a volume of approaches that formulate the problem in a probabilistic framework such as Bayesian networks~\cite{Heckmann2015GraphTD,Lombardo}. 

Neural networks and convolutional models are among more recent approaches for susceptibility mapping. Luo et al.~\shortcite{Luo} and Bui et al.~\shortcite{Bui} use neural networks to assess mine landslide susceptibility and to predict shallow landslide hazards. 
Wang et al.~\shortcite{Wang} did a comparative study on CNNs for landslide susceptibility mapping but their approach does not incorporate any orientational information or aligning filters either. The existing convolutional models are not usually deep and do not use any pooling layers to consider multiple resolutions for feature extraction \cite{nn4}.
Most of these models used in landslide susceptibility mapping are quite simple and do not take into account any orientations. Additionally, their networks do not contain filters that can rotate or capture orientation between landslides. Our proposed CNN architecture is much more sophisticated; it is a fully convolutional network that down-samples images at multiple resolutions and learns filters that can align themselves to the uphill direction. Moreover, our model is trained on geospatial data rather than satellite images.

\begin{figure}[h]
    \centering
    \includegraphics[width=0.55\columnwidth]{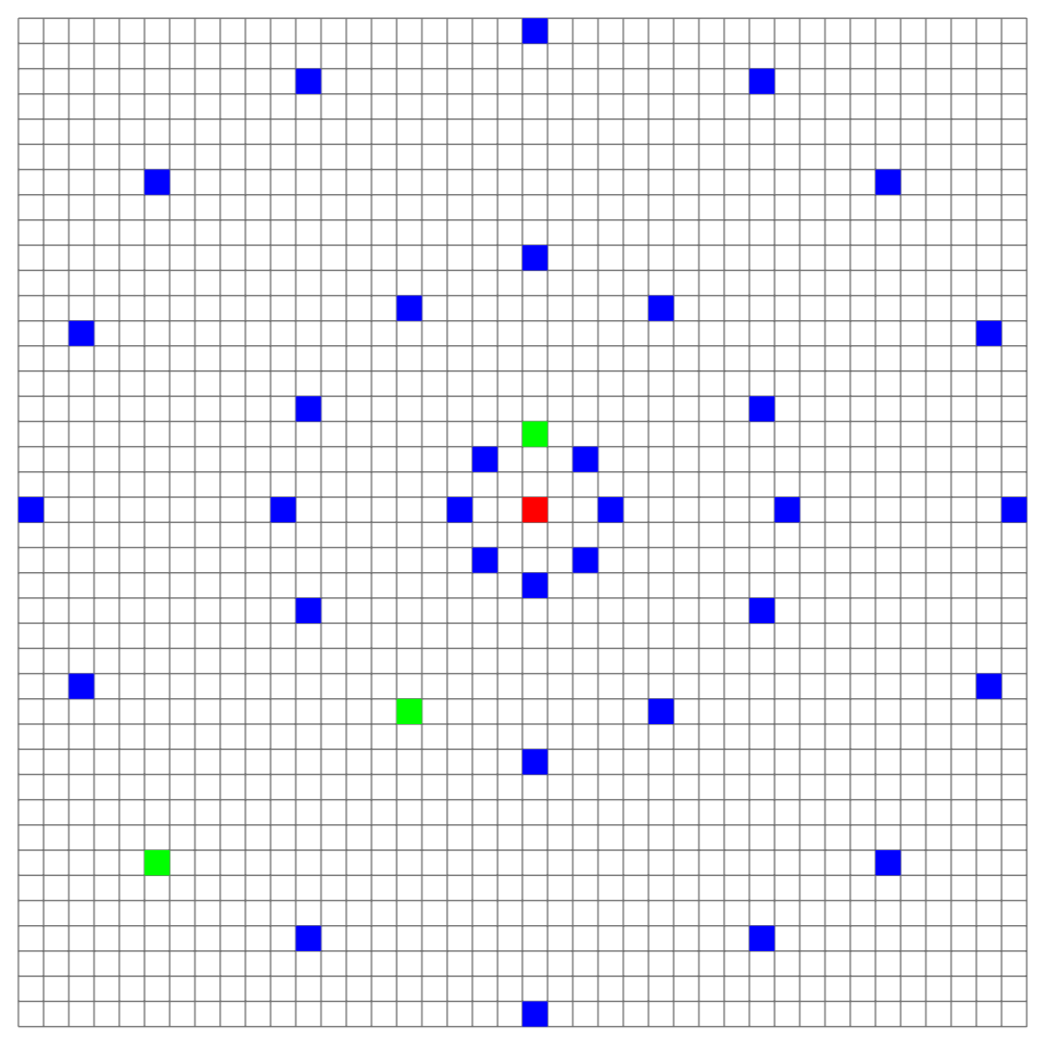}
    \caption{The process of finding aligned features at multiple scales. The red point shows the point of interest where we want to find the uphill directions. Each blue circle shows a set of neighboring points at a specific range. The green point in each circle is the detected point with the highest elevation at that distance, from which the aligned features will be extracted.}
    \label{fig:mask}
\end{figure}

\begin{figure*}[t]
    \centering
    \includegraphics[width=1.8\columnwidth]{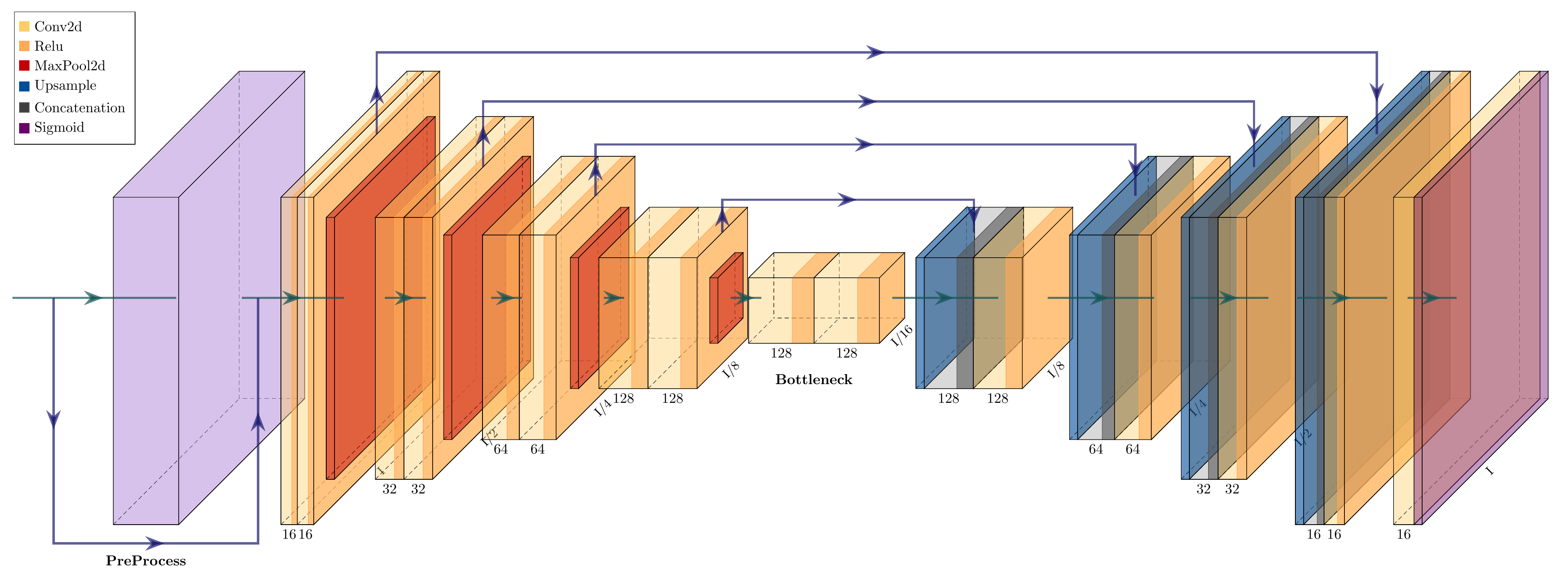}
    \caption{The locally aligned CNN architecture used for predicting landslides. Each conv2d uses a kernel of size 3 with stride 1 and each MaxPool unit uses a kernel of size 2. Upsample units interpolate the image with scale factor of 2 by bi-linear interpolation.}
    \label{fig:CNN}
\end{figure*}

\section{Dataset}

The dataset used for predicting landslides is from Italian open-source databases. The dataset contains both continuous and categorical features in the shape of rasters and vector files respectively. Continuous features including slope and DEM\footnote{Digital Elevation Model} contain out of range values while categorical features such as rock type, land cover, rock age, and rock family, have several no-data points.
To use such data in a CNN, we converted each vector map to a raster after removing invalid data points and out of range values.

As we wanted to propose a baseline framework for this type of problem, we needed to come up with a standard set of features for our categorical data. We chose 44 rock types, 5 land covers, 5 rock families, and 38 rock ages, based on the INSPIRE terminology, as the one-hot encoding for our categorical data. INSPIRE\footnote{Infrastructure for Spatial Information in Europe: \url{https://inspire.ec.europa.eu}} is a European Union directive for standardizing spatial data across countries in Europe. 

Using the INSPIRE terminology, we ended up with 94 standard input features. These features include 44 lithology or rock type features (such as gneiss, mica-schist, granite, and siltstone, etc.), 5 land cover features (agricultural areas, artificial surfaces, forest and semi-natural areas, water bodies, and wetlands), 4 rock family features (metamorphic, sedimentary, plutonic, and volcanic), 38 rock age features (such as paleozoic cycle, cretaceous-jurassic cycle, and average triassic cycle, etc.), and digital elevation model maps, resulting in 92 features. We also use an unknown class for the rock family features along with a slope map. This results in 94 features in total.

We chose Veneto, a region of Italy, since it expands over both mountains and flat zones close to the sea. Each pixel in our prepared dataset has a 10 meters resolution and the images are 21005$\times$19500 pixels resulting in an area with approximately 210 (km) width and 195 (km) height. The ratio of landslides in this region is below 1\% which makes the dataset extremely imbalanced. The landslides in Veneto include both mountainous and less steep areas which are good for training our model. Unfortunately, the landslides do not usually contain information about the date of occurrence.
All of these characteristics make this dataset challenging from the machine learning point of view. The instructions on how to access data are available for other researchers here: \url{https://github.com/ainazHjm/VenetoItaly/}.

\section{Locally Aligned Convolutional Neural Network}
The slope is considered one of the main conditioning factors in predicting landslides. The LLR baseline that we learned also confirms this claim as the slope's weight is among the top 5 learned weights. Traditional CNN filters are oriented vertically in an image, but the important orientation is uphill and downhill for landslides. Based on this, we propose a Locally Aligned CNN model with filters that align themselves according to the uphill direction and extract features alongside that direction. 

For each pixel, we look at three different ranges and choose the highest elevation value at each range, and extract relevant features at those points (refer to Figure \ref{fig:mask}). Because space is at a premium for batch size, we selected a subset of 22 features for this purpose. These features are chosen based on our trained LLR baseline. We chose a feature if the absolute value of its logistic regression weight is 0.2 or greater.

\subsection{Architecture}
Our Locally Aligned CNN architecture consists of a preprocessing module and four layers of down-sampling and up-sampling as in Figure \ref{fig:CNN}. The preprocessing module takes the elevation map along with other input features from the dataset as inputs and outputs 22 aligned features for each looking distance. We use 30, 100, and 300 meters as looking distances in our experiments but it can also be considered a hyper-parameter and be optimized using cross-validation. The preprocessing module outputs 66 aligned features that we further feed into the convolutional network along with the original 94 features. We apply long skip connections between each sampling layer in our LACNN architecture. Each down-sampling layer consists of two convolution layers followed by Relu as non-linearity and a max-pooling layer. Every up-sampling layer includes an up-sampling module to interpolate the data followed by convolutions and Relu. In the end, we apply a Sigmoid function to the output of the model to obtain probabilities.

\subsection{Training}

The rasters in the dataset are too large to fit into a 12 GB memory of a TitanXP GPU when training. Instead, we divide each raster, an input feature, into smaller images of size 500$\times$500, which we call patches. We further feed mini-batches of these patches into our model for training. Since we want to produce a coherent probability map for the whole region, we use patches that overlap each other. For this purpose, we pad each patch with 64 pixels on each side resulting in 628$\times$628 images. This padding number is used to ensure that the overlap between patches is bigger than the receptive field of view of our networks. We partitioned these patches into training, testing, and validation sets.

\begin{figure}
    \centering
    \includegraphics[width=0.65\columnwidth]{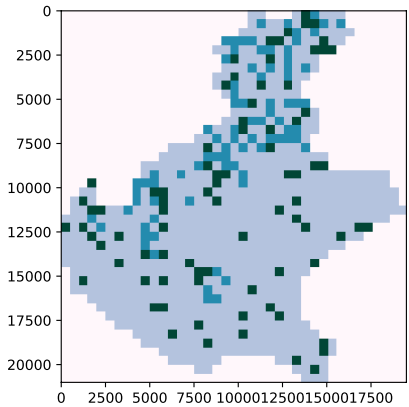}
    \caption{This image shows the data partitioning used for the whole region of Veneto. Grey, blue, and green colors are used to represent train, validation, and test sets respectively. The light background is outside of the region.}
    \label{fig:division}
\end{figure}

We randomly partitioned image patches such that 80\% of the data is used for training, 10\% for testing, and the other 10\% for validation (refer to Figure \ref{fig:division}). We use the negative log-likelihood loss to train our model. However, as the training data is extremely imbalanced, we use oversampling to balance the data to some extent.
Since we want to train our model on patches and preserve the spatial relation between pixels, we oversample patches that have at least one positive label. By oversampling those patches, we are oversampling both landslides and non-landslide pixel points. After doing this, the distribution of landslides stays below 1\%. This oversampling technique can also be seen as a type of data augmentation which provides more training data. We used an oversampling ratio of 5 in our experiments.

\subsection{Hyper-Parameters}
Table \ref{hyper-param} shows the hyper-parameters used for training each of these models. We optimized the learning rate and the optimizer with 5-fold cross-validation for one epoch. The batch size is chosen such that we can fit the maximum number of patches in the memory. The number of epochs is chosen to fully train each model. We validate our models at each epoch and reduce the learning rate if the validation error keeps increasing for $patience$ number of epochs to avoid overfitting. We chose $patience=2$ and $decay=0.001$, which is the L2 regularization lambda, in our experiments. The code is available publicly here: 
\url{https://github.com/ainazHjm/LandslidePrediction/}.

\begin{table}
\caption{Trained Hyper-Parameters. \textit{LR} and \textit{BS} represent the learning rate and the batch size respectively.} \label{hyper-param}
\begin{center}
\begin{tabular}{lcccc}
\toprule
\textbf{MODEL} &\textbf{OPTIMIZER} &\textbf{LR} &\textbf{EPOCHS} &\textbf{BS} \\
\midrule
LLR &Adam &0.125 &10 &15 \\
NN &Adam &0.125 &10 &13 \\
LANN &Adam &0.016 &15 &10 \\
CNN &SGD &0.125 &20 &12 \\
LACNN &Adam &0.001 &30 &9 \\
\bottomrule
\end{tabular}
\end{center}
\end{table}

\begin{figure}[h]
\begin{subfigure}{\columnwidth}
  \centering
  \includegraphics[width=.72\linewidth]{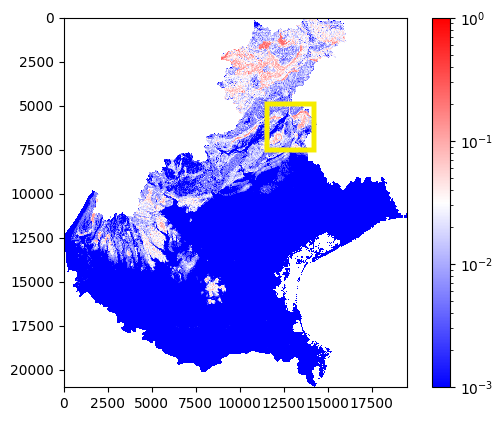}
  \caption{Probability map of the LACNN model on whole Veneto.}
  \label{fig:pdf1}
\end{subfigure}%
\newline
\begin{subfigure}{\columnwidth}
  \centering
  \includegraphics[width=.72\linewidth]{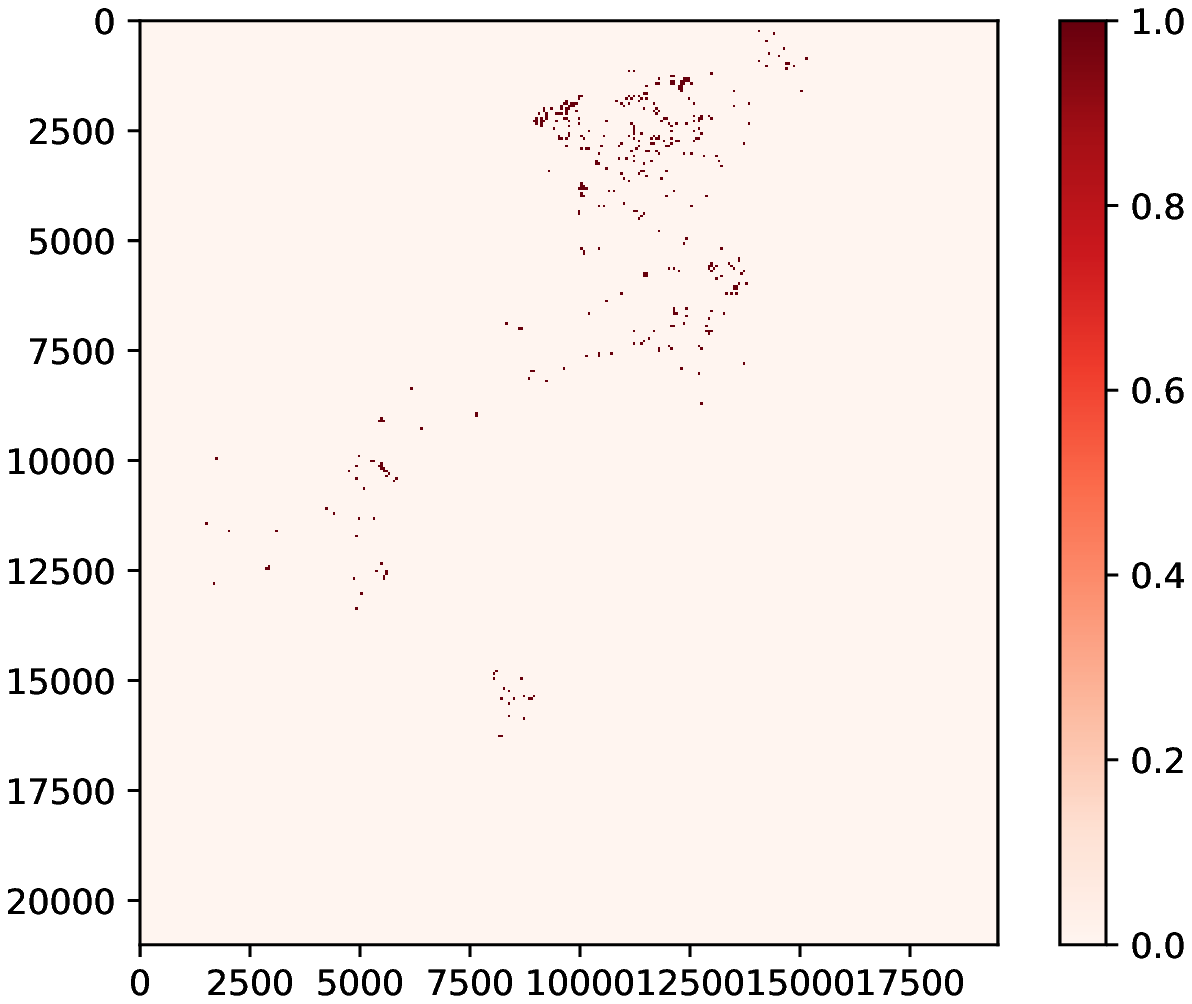}
  \caption{Corresponding ground truth with 21005$\times$19500 resolution.}
  \label{fig:pdf2}
\end{subfigure}
\caption{Red regions correspond to higher probabilities while blue regions are areas with probabilities close to zero. Red polygons represent observed landslides in the area.}
\label{fig:PredvsGt}
\end{figure}

\begin{figure*}[ht]
\begin{subfigure}{.33\textwidth}
  \centering
  \includegraphics[width=.7\linewidth]{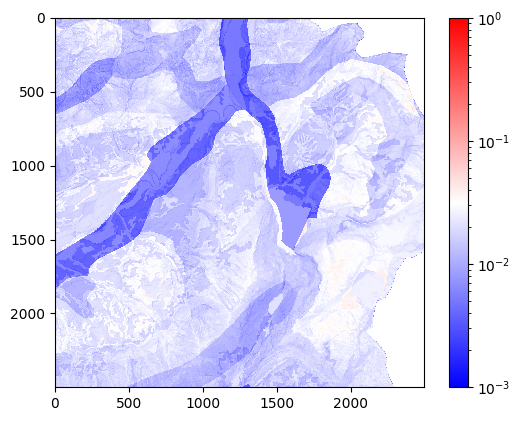}
  \caption{LLR model with 80\% accuracy.}
  \label{fig:sfig1}
\end{subfigure}%
\begin{subfigure}{.33\textwidth}
  \centering
  \includegraphics[width=.7\linewidth]{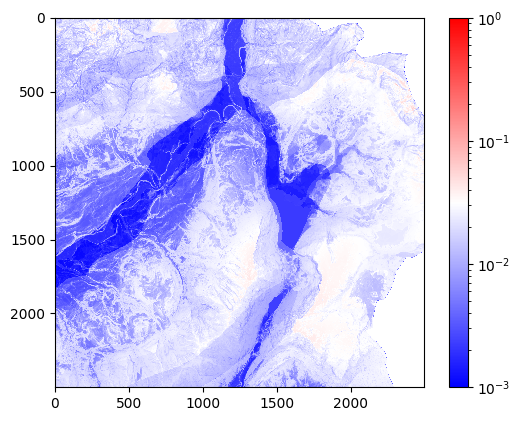}
  \caption{NN model with 83\% accuracy.}
  \label{fig:sfig2}
\end{subfigure}
\begin{subfigure}{.33\textwidth}
  \centering
  \includegraphics[width=.7\linewidth]{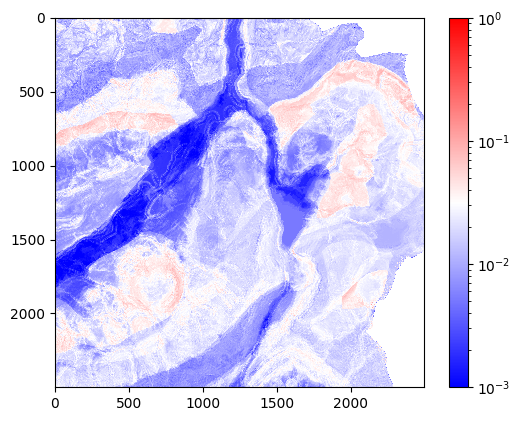}
  \caption{LANN model with 85\% accuracy.}
  \label{fig:sfig3}
\end{subfigure}%
\newline
\begin{subfigure}{.33\textwidth}
  \centering
  \includegraphics[width=.7\linewidth]{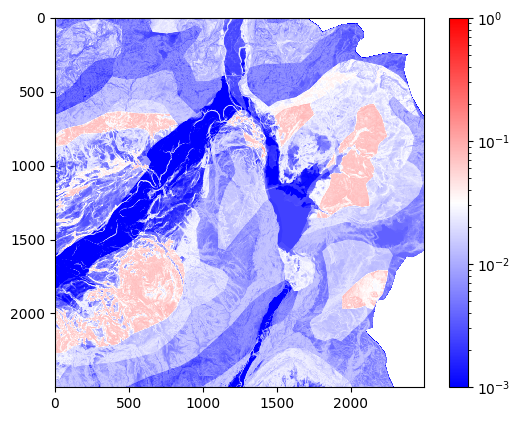}
  \caption{CNN with 85\% accuracy.}
  \label{fig:sfig4}
\end{subfigure}
\begin{subfigure}{.33\textwidth}
  \centering
  \includegraphics[width=.7\linewidth]{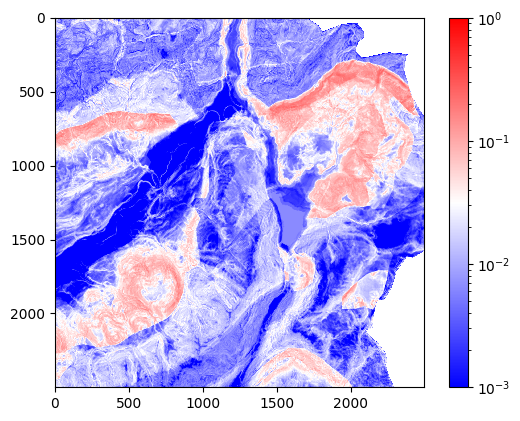}
  \caption{LACNN (ours) with 87\% accuracy.}
  \label{fig:sfig5}
\end{subfigure}
\begin{subfigure}{.33\textwidth}
  \centering
  \includegraphics[width=.7\linewidth]{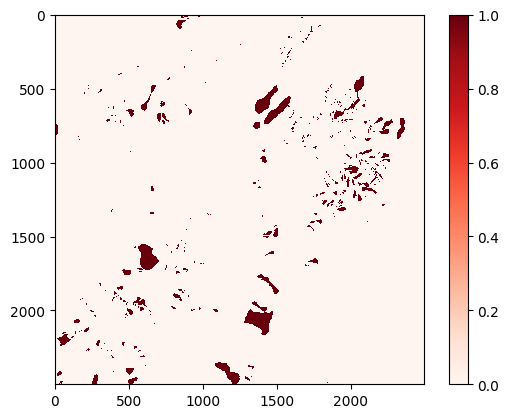}
  \caption{Ground truth.}
  \label{fig:sfig6}
\end{subfigure}%
\caption{Figures \ref{fig:sfig1}-\ref{fig:sfig5} represent the acquired probability maps from different models for a smaller region in Veneto with 2500$\times$2500 resolution and Figure \ref{fig:sfig6} is the corresponding ground truth.}
\label{fig:compare}
\end{figure*}

\subsection{Baselines}\label{sec:calc}
We propose a baseline model called Naive that predicts 0.013 (ratio of landslides in the training set) everywhere in the image. Given the ratio of negative/zero labels, and the ratio of positive/one labels, we can calculate the expected negative log-likelihood error of the Naive baseline on train and test sets. This is approximately equal to 0.069 for the train set and 0.065 for the test set. We compare the test and training errors of our other baselines with the Naive model to make sure that the learned models perform better than Naive (refer to Table \ref{tab:test_error}). 

Our model, LACNN, has two main characteristics: convolutions and feature extraction from the uphill direction. To show the effect of each of these characteristics, we propose a baseline (CNN) that does not pay attention to the uphill direction and only uses convolutions to predict landslides and another baseline (LANN) that does not use any convolutions but extracts features from the uphill direction. We also propose NN, which is a simple neural network that neither uses convolutions nor uphill features, and a linear logistic regression model (LLR) to compare our results to. These models can be looked at as ablation studies to our proposed model (LACNN), which show that locally aligned filters in a convolutional framework are the most effective of all. These baselines are also representative of the current state of the art models that are used in the landslide domain.

\section{Results}
We show the final susceptibility map of our model, LACNN, with the corresponding ground truth in Figures \ref{fig:pdf1} and \ref{fig:pdf2}. This susceptibility map is for the whole region of Veneto. The produced probability map contains many details and can identify areas with high susceptibility around the landslides. Since time scale is not provided for landslides, the output probabilities are for an undefined period, and therefore should only be interpreted as relative scales.

Figure \ref{fig:compare} shows the results of all models for a smaller region of Veneto so that we can compare the output of various models against each other with more details, since the original susceptibility map is too large. The yellow rectangle in Figure \ref{fig:pdf1} shows the corresponding location of this area w.r.t the whole map. It includes both landslide polygons and non-landslide areas and has a variety of terrain. Figures \ref{fig:sfig1}-\ref{fig:sfig5} show that the susceptibility map becomes more detailed as the number of parameters increases and the model gets more complicated. The range of the predicted probabilities is also different between models. More complicated models have a larger variance between their predictions.

Figure \ref{fig:compare} shows how much each model is confident in predicting landslides. Baseline models such as LLR and NN that include a large portion of the state of the art methods, cannot predict landslide with high probability, while more complex models give a better representation of the distribution of data. It is important to note that low confidence does not mean low accuracy. For a small enough threshold, even a LLR baseline can give satisfactory results. However, to produce good susceptibility maps and capture the true distribution of the data, we are interested in richer models with higher confidence.

\subsection{Evaluation Metrics}
\begin{figure}
    \centering
    \includegraphics[width=.9\columnwidth]{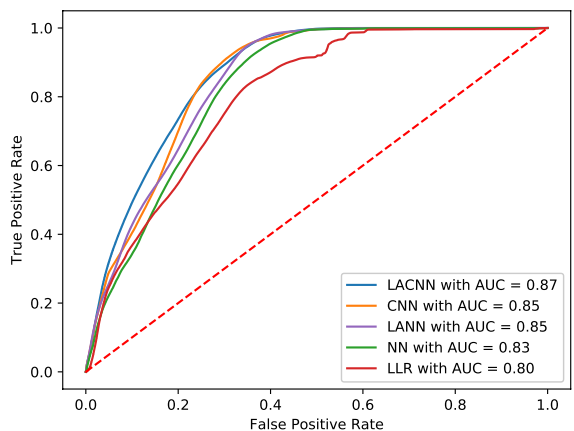}
    \caption{ROC curves from all models on the test set.}
    \label{fig:roc}
\end{figure}

We evaluate our model, LACNN, against other baselines using the Receiver Operating Characteristic (ROC) curve on the test set and show that it achieves the best results (by a margin of 2\% to 7\%) among all models, as shown in Figure \ref{fig:roc}. Note that the LANN model achieves similar performance to the CNN model. This is interesting since this model does not use any convolutions for predictions and only looks at the uphill direction at three different distances (30, 100, and 300 meters), suggesting that alignment plays a significant role in predicting landslides. 

We further assess our model by computing the negative log-likelihood error on both the training and the test sets. Table \ref{tab:test_error} illustrates the results of all baselines. The LACNN model obtains the lowest errors on both train and test sets, improving the baselines by a margin of 2\% to 15\%.

\begin{table}
    \caption{Negative Log Likelihood Loss} 
    \label{tab:test_error}
        \begin{center}
        \begin{tabular}[width=\textwidth]{lccc}
            \toprule
            \textbf{METHOD} &\textbf{TEST ERR} &\textbf{TRAIN ERR} &\textbf{AUC}\\
            \midrule
            Naive &0.065 &0.069 &0.50\\
            LLR &0.055 &0.057 &0.80\\
            NN &0.052 &0.055 &0.83\\
            LANN &0.048 &0.052 &0.85\\
            CNN &0.047 &0.051 &0.85\\
            LACNN &\textbf{0.046} &\textbf{0.050} &\textbf{0.87}\\
            \bottomrule
        \end{tabular}
\end{center}
\end{table}

\section{Conclusion}

Landslides are the movement of ground under the force of gravity. They are common phenomena that can cause significant casualties. There have been many approaches to produce susceptibility maps to reduce the impact of landslides including expert-based, physics-based, and statistical methods. All of these methods have their flaws and lack a standard set of features. We provide a standardized open-source dataset with the same terminology as INSPIRE so that anyone who uses the INSPIRE terminology can compare their results to our proposed baselines. We also propose a novel statistical approach for predicting landslides using machine learning. We introduce a deep convolutional model, called LACNN, that can follow the ground surface and align itself with the ground contour lines to extract relevant features. We evaluate our model by ROC curves and negative log-likelihood error and show that it can achieve the best results on the test set among all the baselines. Our results suggest that this type of statistical approach is effective for generating susceptibility maps which in turn has the potential to alleviate human and financial losses caused by landslides.

\section*{Acknowledgments}
We thank Minerva Intelligence Inc., MITACS, NSERC, and Compute Canada who provided resources and supported this research financially. We thank the staff at Minerva, especially Clinton Smyth, CTO, for providing insight and expertise that greatly assisted this research. We would also like to show our gratitude to Blake Boyko, GIS analyst at Minerva, for helping us with gathering and processing the data.

\bibliographystyle{named}
\bibliography{ijcai20}

\end{document}